\documentclass[sigplan,screen,nonacm]{acmart}
\AtBeginDocument{%
  }
\usepackage{braket}
\usepackage{amsmath,amsfonts}
\usepackage{tikz}
\usepackage{quantikz}
\usepackage{graphicx}
\usepackage{subcaption}
\usepackage{caption}
\usepackage{url}

\def\ie{{\it i.e.}\hspace{0.1pc}}
\def\eg{{\it e.g.}\hspace{0.1pc}}
\def\etal{{\it et al.}\hspace{0.1pc}}

\setcopyright{acmlicensed}
\copyrightyear{2018}
\acmYear{2018}

\acmDOI{XXXXXXX.XXXXXXX}
\acmConference[Conference acronym 'XX]{Make sure to enter the correct
  conference title from your rights confirmation email}{June 03--05,
  2018}{Woodstock, NY}
\acmISBN{978-1-4503-XXXX-X/2018/06}

\begin{document}

\title{Quantum Graph Transformer for NLP Sentiment Classification}


\author{Shamminuj Aktar}
\affiliation{
    \institution{CCS-3 Information Sciences \\ Los Alamos National Laboratory} 
    \city{Los Alamos, NM}
    \country{USA}
    }
\email{saktar@lanl.gov}

\author{Andreas Bärtschi}
\affiliation{
    \institution{CCS-3 Information Sciences \\ Los Alamos National Laboratory} 
    \city{Los Alamos, NM}
    \country{USA}
    }
\email{baertschi@lanl.gov}

\author{Abdel-Hameed A.\ Badawy}
\affiliation{
    \institution{Klipsch School of Electrical and Computer Engineering \\ New Mexico State University} 
    \city{Las Cruces, NM}
    \country{USA}
    }
\email{badawy@nmsu.edu}

\author{Stephan Eidenbenz}
\affiliation{
    \institution{CCS-3 Information Sciences \\ Los Alamos National Laboratory} 
    \city{Los Alamos, NM}
    \country{USA}
    }
\email{eidenben@lanl.gov}
\begin{abstract}
  Quantum machine learning is a promising direction for building more efficient and expressive models, particularly in domains where understanding complex, structured data is critical. We present the Quantum Graph Transformer (QGT), a hybrid graph-based architecture that integrates a quantum self-attention mechanism into the message-passing framework for structured language modeling. The attention mechanism is implemented using parameterized quantum circuits (PQCs), which enable the model to capture rich contextual relationships while significantly reducing the number of trainable parameters compared to classical attention mechanisms. We evaluate QGT on five sentiment classification benchmarks. Experimental results show that QGT consistently achieves higher or comparable accuracy than existing quantum natural language processing (QNLP) models, including both attention-based and non-attention-based approaches. When compared with an equivalent classical graph transformer, QGT yields an average accuracy improvement of 5.42\% on real-world datasets and 4.76\% on synthetic datasets. Additionally, QGT demonstrates improved sample efficiency, requiring nearly 50\% fewer labeled samples to reach comparable performance on the Yelp dataset. 
  These results highlight the potential of graph-based QNLP techniques for advancing efficient and scalable language understanding.
\end{abstract}




\maketitle

\section{Introduction}
    Artificial Intelligence (AI) has become an integral part of modern technological advancement, influencing domains such as healthcare, finance, scientific discovery, and communication~\cite{russell2016artificial}. Natural Language Processing (NLP) plays a foundational role in AI by enabling machines to understand, interpret, and generate human language~\cite{priyadarshini2020brief}. This capability allows NLP to power a variety of applications, from sentiment analysis to machine translation, conversational agents, and content moderation~\cite{moslem2023adaptive,tan2023survey}. Recent advancements in NLP models, such as GPT-4, have significantly improved machines’ ability to generate coherent and context-aware text~\cite{openai2023gpt}. Despite their success, current NLP models still face substantial limitations. Training and inference for state-of-the-art models require large amounts of labeled data, access to high-performance computing (HPC) infrastructure, and the optimization of billions of parameters~\cite{ding2023hpc}.

    Quantum Machine Learning (QML) is a compelling framework for overcoming some of the inefficiencies associated with classical machine learning (ML) models~\cite{biamonte2017quantum, cerezo2022challenges}. QML adopts a hybrid quantum-classical strategy, in which data is encoded into a high-dimensional Hilbert space and learning is performed using Parameterized Quantum Circuits (PQCs), which act as trainable layers analogous to neural network layers in classical architectures. These circuits leverage quantum properties such as superposition and entanglement to explore complex functions more efficiently~\cite{schuld2019quantum,havlivcek2019supervised}. 
    In the context of NLP, where understanding contextual relationships between tokens is essential, QML offers a promising direction for developing more efficient and expressive language models. Quantum-enhanced NLP (QNLP) models have the potential to achieve quantum advantages in sample efficiency, model accuracy, convergence speed, and generalization capability. 

    Recent efforts in QNLP have explored various approaches to integrate quantum principles into language modeling. Early models include the quantum bag-of-words approach, quantum support vector machines (QSVMs) and variational quantum classifiers employing amplitude encoding~\cite{lorenz2023qnlp,alexander2022quantum}. More recent developments have focused on hybrid architectures such as quantum-enhanced LSTMs~\cite{stein2023applying}, self-attention based models~\cite{li2024quantum, chen2025quantum, zhang2024light}, and ensemble based models like LEXIQL~\cite{silver2024lexiql}. These approaches have shown promising results, indicating the potential of QNLP as a viable and scalable alternative to classical NLP models. 

    In classical ML, graph-based models have demonstrated strong performance in capturing the structure of complex data, including knowledge graphs, citation networks, and dependency trees in NLP~\cite{wu2020comprehensive, liu2022graph}. Graph neural networks (GNNs) represent inputs as graphs, where nodes represent entities (such as tokens in NLP) and edges capture syntactic or semantic relationships~\cite{scarselli2008graph}. This graph-based representation enables models to effectively capture both hierarchical and contextual relationships within the data.
    
    In this work, we propose the Quantum Graph Transformer (QGT), a novel hybrid quantum-classical model for sentiment classification. QGT represents each sentence as a fully connected graph of tokens and applies a modified quantum transformer convolution (QTransformerConv) layer to propagate information across the nodes. The self-attention mechanism in the QTransformerConv layer leverages PQCs to generate the query and key vectors necessary for computing attention scores between neighboring nodes. Specifically, each token node is encoded into a quantum state, and then PQCs parameterized by learnable weights are applied to extract the query and key representations. The resulting attention scores guide the flow of information across the graph, allowing the model to capture contextual dependencies among tokens in a quantum-enhanced feature space. We evaluate the QGT on five benchmark sentiment classification datasets and demonstrate that QGT efficiently learns contextual representations of token embeddings. We also compare the model performance on those benchmark datasets with other QNLP models and a classical graph transformer model.

    Our main contribution are as follows:
    \begin{enumerate}
        \item We introduce a hybrid graph-based architecture that integrates a quantum self-attention mechanism into the message-passing framework for structured language modeling. Our design leverages PQCs to implement self-attention, requiring significantly fewer trainable parameters than traditional classical attention mechanisms while maintaining high expressibility.
        \item We evaluate the proposed QGT model on five sentiment classification benchmarks, including three real-world datasets \ie Yelp, IMDB, and Amazon~\cite{kotzias2015group} and two synthetic datasets \ie MC and RP~\cite{lorenz2023qnlp}. The QGT model consistently learns meaningful contextual representations and accurately predicts sentiment labels across all datasets. It outperforms or matches the accuracy of existing attention-based QNLP models and other QNLP baselines.
        \item To further assess model performance, we compare QGT with an equivalent classical graph transformer model. Our results show that QGT achieves significantly higher accuracy, with an average improvement of 5.42\% on three real-world datasets and 4.76\% on two synthetic datasets.
        \item Additionally, QGT demonstrates enhanced sample efficiency, requiring nearly 50\% fewer training samples to achieve comparable accuracy on the Yelp dataset.
    \end{enumerate}

    \begin{figure*}[t!]  
        \centering
        
        \includegraphics[width=0.99\textwidth]{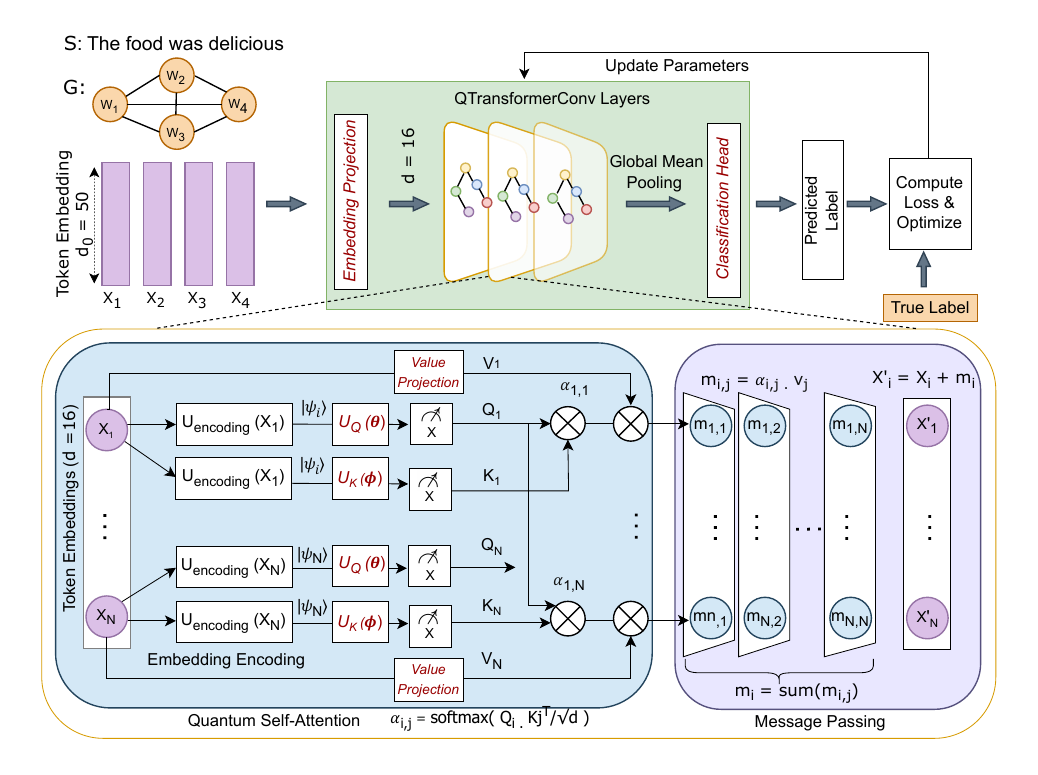}  
        \captionsetup{skip=9pt} 
        \caption{Overview of the Quantum Graph Transformer (QGT) architecture. Each input sentence $\mathcal{S}$ is tokenized and represented as a fully connected graph $G$. Tokens are initialized using GloVe embeddings ($d_0 = 50$) and projected to $\mathbf{x}_i \in \mathbb{R}^d$ with $d = 16$, then processed through stacked QTransformerConv layers and aggregated via global mean pooling. In each QTransformerConv layer, node features $\mathbf{x}_i$ are encoded into quantum states $\ket{\psi_i}$ and processed using $U_Q(\boldsymbol{\theta})$ and $U_K(\boldsymbol{\phi})$ to extract query and key vectors. These vectors are used in quantum self-attention to compute attention scores $\alpha_{i,j}$, guiding message passing to update node features. The aggregated features are passed through a classification head for label prediction, and model parameters (\textit{marked in red}) are updated via backpropagation to minimize the loss.}

        \label{fig:qgt-model}
    \end{figure*}

\section{Background \& Motivation}
    \subsection{Natural Language Processing Tasks}
    Natural Language Processing (NLP) is a specialized domain within AI that aims to bridge the gap between human communication and machine understanding by enabling computational systems to process and generate natural language. Core NLP tasks include text classification, named entity recognition, question answering, and machine translation. These tasks are unified by the need to model complex syntactic and semantic relationships within and between sequences of text. In the sentiment classification task, the objective is to predict the underlying emotion or opinion expressed in a given sentence or document, typically as a categorical label such as positive, negative, or neutral. Sentiment classification is a widely studied problem due to its practical relevance in areas such as customer feedback analysis, user experience assessment, and automated review aggregation.
    \subsection{Classical NLP Models}
    In the early stages of natural language processing, models primarily relied on simple representations such as Bag-of-Words and N-gram features, which treated text as unordered collections of word counts without accounting for syntax or semantics~\cite{harris1954distributional, brown1992class}. Subsequently, more linguistically informed approaches emerged, utilizing grammar-aware structures such as dependency trees and part-of-speech tags to capture syntactic relationships between words in a sentence. Although these structured models offered improved linguistic insight, they often struggled to generalize effectively across diverse language contexts. A major advancement came with the introduction of word embeddings such as Word2Vec~\cite{church2017word2vec} and GloVe~\cite{pennington2014glove}, which represent words as continuous vectors in a dense embedding space. These embeddings capture semantic similarity based on word co-occurrence patterns, enabling models to generalize better across related linguistic constructs. Models such as recurrent neural networks (RNNs)~\cite{tang2015document}, long short-term memory networks (LSTMs)~\cite{hochreiter1997long}, and gated recurrent units (GRUs)~\cite{cho2014learning} leveraged these embeddings to process sequential data and learn contextual representations of text. However, these models struggled to capture long-range dependencies due to issues like vanishing gradients and limited memory capacity. 

    Another breakthrough in NLP came with the introduction of the attention mechanism, which enables models to dynamically focus on relevant parts of the input sequence. This concept was central to the development of the Transformer architecture~\cite{vaswani2017attention}, which entirely replaces recurrence with self-attention layers. Models such as BERT~\cite{devlin2019bert}, GPT~\cite{floridi2020gpt}, and RoBERTa~\cite{liu2019roberta} achieved state-of-the-art results across a wide range of NLP tasks. These models typically require large amounts of labeled data, substantial computational resources, and still face challenges in generalizing to unseen linguistic patterns.

    \subsection{Existing Quantum NLP Models}
    Quantum Machine Learning (QML) has been extensively studied in recent years due to its potential to surpass the capabilities of classical machine learning and demonstrate quantum advantage. QML typically employs a hybrid quantum--classical approach, where PQCs are used as trainable quantum layers. A PQC is represented as a unitary transformation \( U(\boldsymbol{\lambda}) \) acting on an input quantum state \( \ket{\psi_{\text{in}}} \), where \( \boldsymbol{\lambda} = \{\lambda_1, \lambda_2, \ldots, \lambda_p\} \) are tunable real-valued parameters. The resulting output state is given by $\ket{\psi_{\text{out}}} = U(\boldsymbol{\lambda}) \ket{\psi_{\text{in}}}$. In the QML framework, classical input data is first encoded into a quantum state, and the PQC is subsequently trained to learn the optimal parameters that minimize the loss function for the target learning task.

    Recently, there has been growing interest in QNLP models as a promising alternative to classical NLP models. Hybrid quantum-enhanced NLP models aim to encode natural language into quantum states and optimize parameters to learn semantic relationships between words. Since quantum states reside in high-dimensional Hilbert space, they offer superior expressive power, even with relatively few parameters. The categorical compositional distributional (DisCoCat) model combines distributional semantics with grammatical structure and maps naturally onto quantum tensor product spaces \cite{coecke2010mathematical}. However, the challenge of accurately extracting syntactic structure from sentence inputs limits the adaptability of the DisCoCat model, especially when handling the variability and ambiguity of natural language. Other approaches include quantum-enhanced bag-of-words (BoW)~ and N-gram models, which encode word occurrences or co-occurrence patterns into quantum states~\cite{alexander2022quantum}. These models have shown promising results on synthetic datasets, but their practical implementation and performance are often constrained by the qubit requirements, limiting their scalability and real-world applicability. In addition, Quantum LSTM~\cite{stein2023applying} and Quantum SVM~\cite{alexander2022quantum} models have been proposed to address sentiment classification tasks similar to classical NLP models. Recent efforts have focused on developing quantum-enhanced self-attention mechanisms. These models demonstrated better learning on real-world benchmark datasets~\cite{li2024quantum, zhang2024light, chen2025quantum}. Additionally, a recent work by Silver \etal introduced an ensemble-based technique that employs an incremental data injection approach to improve generalization in quantum NLP models for sentiment classification tasks \cite{silver2024lexiql}.
        
    \subsection{Classical Graph Transformers}
    Classical graph transformer are a class of graph neural network models that incorporate the self-attention mechanism from the original transformer architecture into graph structured data~\cite{yun2019graph,  vaswani2017attention}. Specifically, graph transformer replaces fixed  graph convolution with learned attention-based message passing. This allows each node to attend to a subset or all other nodes based on learned attention scores. As a result, graph transformers can effectively capture both local and long-range dependencies in the underlying data. Given a undirected, unweighted graph $G = (V, E)$ with node features $\left \{\mathbf{x}_i \in \mathbb{R}^d\right \}_{i \in V}$, each node undergoes a feature transformation through three learned linear projections to obtain query, key and value representations:
    \begin{align}
        \mathbf{Q}_i = \mathbf{W}_Q \cdot \mathbf{x}_i, \: \mathbf{K}_j = \mathbf{W}_K \cdot \mathbf{x}_j,  \: \mathbf{V}_j = \mathbf{W}_V \cdot \mathbf{x}_j, 
    \end{align}
    Here $\mathbf{W}_Q \text{,}\: \mathbf{W}_K \text{,}\: \mathbf{W}_V \in \mathbb{R}^d$ are shared trainable parameters. The attention scores $\alpha_{i,j}$ is computed between target node $i$ and its neighbor $j$ using a scaled dot-product between the query and key vectors: 
    \begin{align}
        e_{i,j} = \frac{\mathbf{Q}_i \cdot \mathbf{K}_j^T}{\sqrt{d}}
        \label{classical-attn}
    \end{align}
    To ensure these scores are comparable across neighbors, they are normalized using the softmax function:
    \begin{align}
        \alpha_{i,j} = \frac{\exp(e_{i,j})}{\sum_{m \in \mathcal{N}(i)} \exp(e_{i,m})}
        \label{classical-attn-normalized}
    \end{align}
    where $\mathcal{N}(i)$ denotes the set of neighbors of node $i$. The resulting attention coefficients $\alpha_{i,j}$ reflects the relative importance or relevance of each neighbor $j$ when updating node $i$'s feature vector. The final output feature for node $i$ is computed as a weighted sum of the value vectors of its neighbors:
    \begin{align}
        \mathbf{x}'_i = \mathbf{x}_i + \sum_{j \in \mathcal{N}(i)} \alpha_{i,j} \cdot \mathbf{V}_j
    \end{align}
    This formulation enables each node to selectively focus on different parts of its neighborhood, allowing the model to dynamically encode structural information based on relevancy score. Multiple layers of transformer-based convolution can be stacked to progressively refine node representations and capture higher-order dependencies across the graph. 

    In today's AI-driven world, where improving natural language understanding is becoming more crucial, quantum-enhanced self-attention models have demonstrated significant potential in advancing NLP tasks. Inspired by this, we propose the Quantum Graph Transformer (QGT) model for sentiment classification for capturing deeper context through quantum-enhanced representations.
    
    
    \begin{figure*}[t!]
        \centering
        \begin{subfigure}[b]{0.47\textwidth}
            \centering
            \begin{quantikz}[column sep=0.13cm, row sep=0.2cm]
                \lstick{$0\colon\ket{0}$} & \gate{R_y\left(\mathbf{x}_i^0 + \mathbf{x}_i^1 +\mathbf{x}_i^2 +\mathbf{x}_i^3 \right)} & \ctrl{1}              & \qw         & \qw         & \targ{}     & \qw\rstick[4]{$\ket{\psi_i}$} \\
                \lstick{$1\colon\ket{0}$} & \gate{R_y\left(\mathbf{x}_i^4 + \mathbf{x}_i^5 + \mathbf{x}_i^6 + \mathbf{x}_i^7\right)} & \targ{}               & \ctrl{1}    & \qw         & \qw         & \qw \\
                \lstick{$2\colon\ket{0}$} & \gate{R_y\left(\mathbf{x}_i^8 + \mathbf{x}_i^9 + \mathbf{x}_i^{10} + \mathbf{x}_i^{11}\right)} & \qw                   & \targ{}     & \ctrl{1}    & \qw         & \qw \\
                \lstick{$3\colon\ket{0}$} & \gate{R_y\left(\mathbf{x}_i^{12} + \mathbf{x}_i^{13} + \mathbf{x}_i^{14} + \mathbf{x}_i^{15}\right)} & \qw                   & \qw         & \targ{}     & \ctrl{-3}   & \qw
            \end{quantikz}
        \end{subfigure}%
        \hfill
        \begin{subfigure}[b]{0.53\textwidth}
            \centering
            \begin{quantikz}[column sep=0.13cm, row sep=0.2cm]
                \lstick[4]{ } & \gate{R_x(\boldsymbol{\theta}_0)} & \gate{R_y(\boldsymbol{\theta}_4)} & \gate{R_z(\boldsymbol{\theta}_8)} & \ctrl{1}              & \qw         & \qw         & \targ{}     & \gate{R_y(\boldsymbol{\theta}_{12})} & \gate[wires=4]{QFT} & \qw \\
                \qw & \gate{R_x(\boldsymbol{\theta}_1)} & \gate{R_y(\boldsymbol{\theta}_5)} & \gate{R_z(\boldsymbol{\theta}_9)} & \targ{}               & \ctrl{1}    & \qw         & \qw         & \gate{R_y(\boldsymbol{\theta}_{13})} &                         & \qw \\
                \qw & \gate{R_x(\boldsymbol{\theta}_2)} & \gate{R_y(\boldsymbol{\theta}_6)} & \gate{R_z(\boldsymbol{\theta}_{10})} & \qw                   & \targ{}     & \ctrl{1}    & \qw         & \gate{R_y(\boldsymbol{\theta}_{14})} &                         & \qw \\
                \qw & \gate{R_x(\boldsymbol{\theta}_3)} & \gate{R_y(\boldsymbol{\theta}_7)} & \gate{R_z(\boldsymbol{\theta}_{11})} & \qw                   & \qw         & \targ{}     & \ctrl{-3}   & \gate{R_y(\boldsymbol{\theta}_{15})} &                         & \qw
            \end{quantikz}
    \end{subfigure}
   
    \caption{({\textit{Left}}) Quantum embedding circuit $U_{\text{encoding}}(\mathbf{x}_i)$ maps a token embedding vector $\mathbf{x}_i \in \mathbb{R}^{d=16}$ to a 4-qubit quantum state $\ket{\psi_i}$ using grouped $R_y$ rotations, where each rotation angle is given by $\smash{\sum_{j=4k}^{4k+3} \mathbf{x}_i^j}$ for qubits $k = 0, 1, 2, 3$. The circuit is then followed by a ring-style CNOT entanglement layer. ({\textit{Right}}) Parameterized quantum circuit $U_Q(\boldsymbol{\theta})$ used for query vector generation consists of $R_x$, $R_y$ and $R_z$ layers per qubit, a ring-style CNOT entanglement layer, a second $R_y$ layer, and a Quantum Fourier Transform (QFT)~\cite{weinstein2001implementation} at the end. The combined circuit $U_Q(\boldsymbol{\theta}) \cdot U_{\text{encoding}}(\mathbf{x}_i)$ is applied, and measurement yields the query vector $\mathbf{Q}_i$. Similarly, key vector $\mathbf{K}_i$ is generated using parameters $\boldsymbol{\phi}$.}
    \label{fig:qgt-circuits}
    \end{figure*}
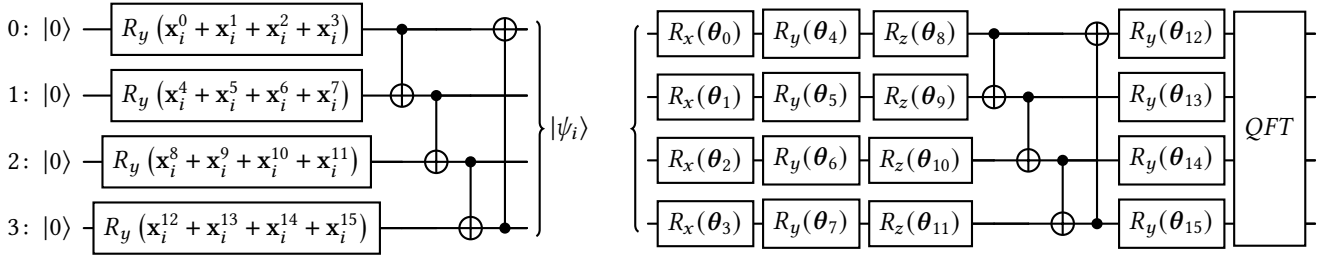
\section{Quantum Graph Transformer Model}
\label{sec:QGT-Model}
    Figure~\ref{fig:qgt-model} illustrates the overall architecture of the proposed QGT model. The hybird QGT model intergrates quantum self-attention into a graph-based nerual architecture for sentiment classification. The architecture of the model consists of the following key components: 
    
    \subsection{Sentence to Graph Construction}
    Each input sentence \( \mathcal{S} \in \mathcal{D} \), where \( \mathcal{D} \) denotes the dataset and \( \mathcal{S} \) represents one sentence, is first tokenized into a sequence of tokens. This sequence is denoted as \( \mathcal{S} = \{ w_1, w_2, \\\dots, w_N \} \), where each \( w_i \) represents a token and \( N \) denotes the total number of tokens in the sentence after tokenization. Based on this tokenized sequence, we construct a graph $G = (V, E)$, where $V$ represents the tokens treated as nodes and $E$ corresponds to the edge connections between tokens. In the QGT model, we consider a complete graph structure in which every node is connected to every other node, i.e., $(i, j) \in E $ for all $i \neq j$. The dense connectivity allows each token to directly exchange information with all other tokens during message passing. This is particularly essential for capturing long-range dependencies and global context required in sentiment analysis tasks. Alternatively, for efficiency and local context modeling, a $k$-nearest neighbor (k-NN) graph can be employed, where each node is connected only to its $k$ immediate neighbors in the token sequence. This sparsity can significantly reduce computational overhead, particularly for shorter sentences (\eg tweets, chat messages, or search queries) where full connectivity may not be necessary.
    
    \subsection{QTransformerConv: Embedding Encoding}
        Each node $i \in V$ (corresponding to token $w_i$) is initially mapped to a pretrained GloVe embedding vector~\cite{pennington2014glove} $em_{i} \in \mathbb{R}^{50}$, capturing semantic and syntactic information of the token.  First, we project GloVe embeddings to a lower dimensional representation $\mathbf{x}_i \in \mathbb{R}^{d}$ using a linear layer. Here, $d=16$ is the reduced dimension of each node feature vector. The dimensionality-reduced input embedding vectors can be efficiently encoded into quantum circuits, thereby requiring fewer qubits. The projected feature $\mathbf{x}_i$ is then encoded into a quantum state $\ket{\psi_i}$ using the quantum encoding circuit $U_{\text{encoding}}(\mathbf{x}_i)$ shown in Figure~\ref{fig:qgt-circuits} (left). The circuit utilizes 
        $n = \sqrt{d}$ qubits, 
        where each qubit is initialized by applying $n$ number of $R_y$ gates sequentially corresponding to the components of $\mathbf{x}_i$. Following these $R_y$ rotations, entanglement is introduced through a series of CNOT gates arranged in a ring pattern, connecting each qubit to its neighboring qubit. The encoding circuit can be represented like this
        \begin{align}
           U_{\text{encoding}}(\mathbf{x}_i) = \left( \prod_{k=0}^{n-1} \text{CNOT}_{k,\,k+1} \right) \cdot  \left( \bigotimes_{k=0}^{n-1} \left( \prod_{j=n k}^{n k+n-1} R_y(\mathbf{x}_i^j) \right) \right)
        \end{align}
        Since each qubit undergoes $n$ sequential $R_y$ rotations, these rotations can be merged into a single $R_y$ gate with a cumulative rotation angle. This simplification is expressed as:
        \begin{align} 
            \prod_{j=nk}^{nk+n-1} R_y(\mathbf{x}_i^j) \equiv R_y\left( \sum_{j=nk}^{nk+n-1} \mathbf{x}_i^j \right) 
        \end{align}
        Applying this encoding circuit to the initial state $\ket{0}^{\otimes n}$ yields the encoded quantum state:
       \begin{align}
            U_{\text{encoding}}(\mathbf{x}_i) = 
            \left( \prod_{k=0}^{n-1} \text{CNOT}_{k,\, k+1} \right) 
            \cdot  
            \left( \bigotimes_{k=0}^{n-1} R_y\left( \sum_{j=nk}^{nk+n-1}\mathbf{x}_i^j \right) \right)
        \end{align}

        \begin{align} 
            \ket{\psi_i} = U_{\text{encoding}}(\mathbf{x}_i) \ket{0}^{\otimes n} 
        \end{align}

        \begin{figure*}[t]
            \centering
            \vspace{-5px}
            \begin{subfigure}[b]{0.33\textwidth}
                \includegraphics[width=\linewidth]{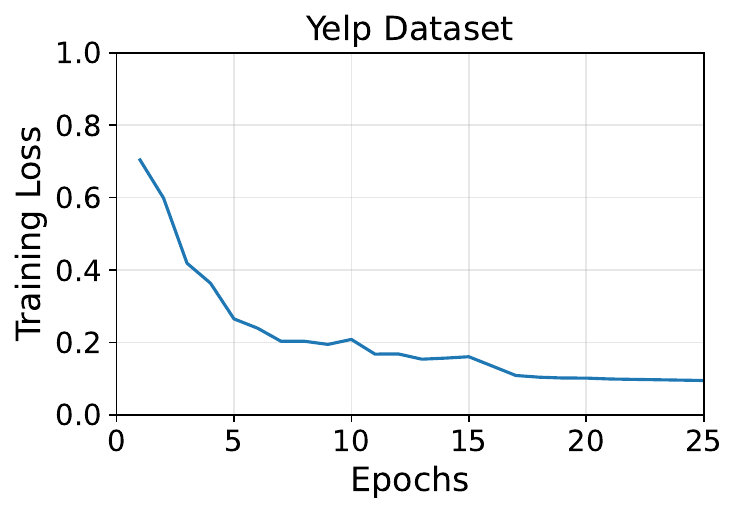}
                
            \end{subfigure}
            \hfill
            \begin{subfigure}[b]{0.33\textwidth}
                \includegraphics[width=\linewidth]{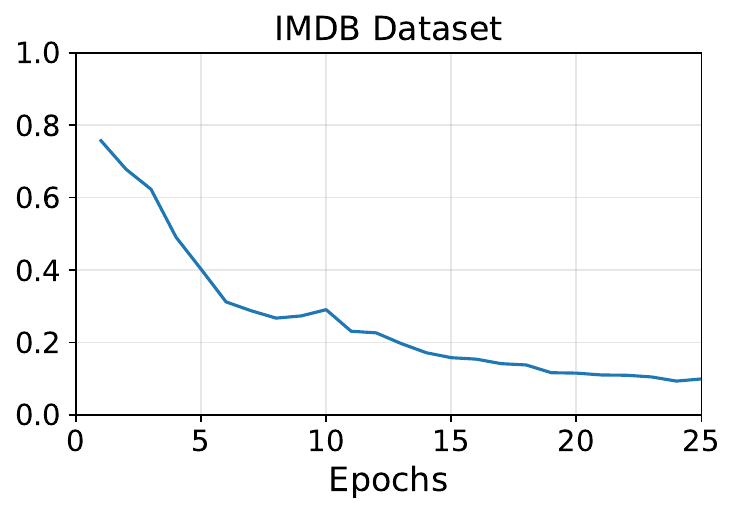}
               
            \end{subfigure}
            \hfill
            \begin{subfigure}[b]{0.33\textwidth}
                \includegraphics[width=\linewidth]{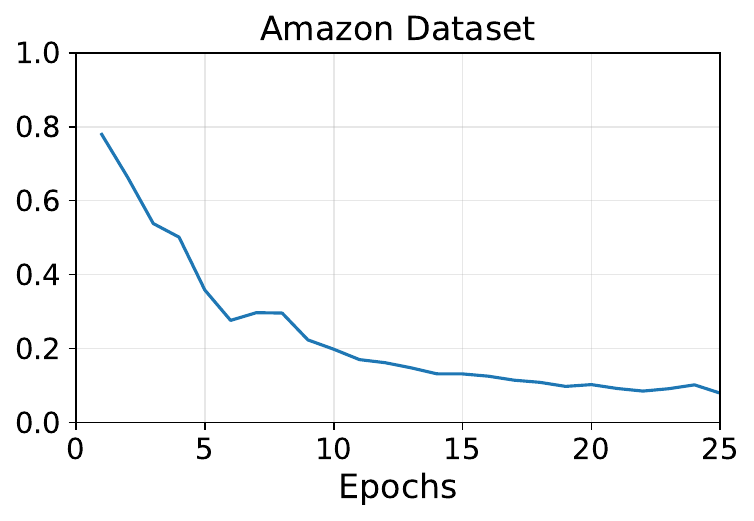}
            \end{subfigure}
            \captionsetup{skip=2pt} 
            \caption{Training loss curves for the Yelp (\textit{left}), IMDB (\textit{middle}), and Amazon (\textit{right}) datasets in the sentiment classification task using the proposed QGT model. All three datasets show consistent loss reduction /learning over training epochs.} 
            \label{fig:model-learning-plots}
        \end{figure*}
        \begin{figure*}[t]
        \vspace{-5px}
            \centering
            \begin{subfigure}[b]{0.33\textwidth}
                \includegraphics[width=\linewidth]{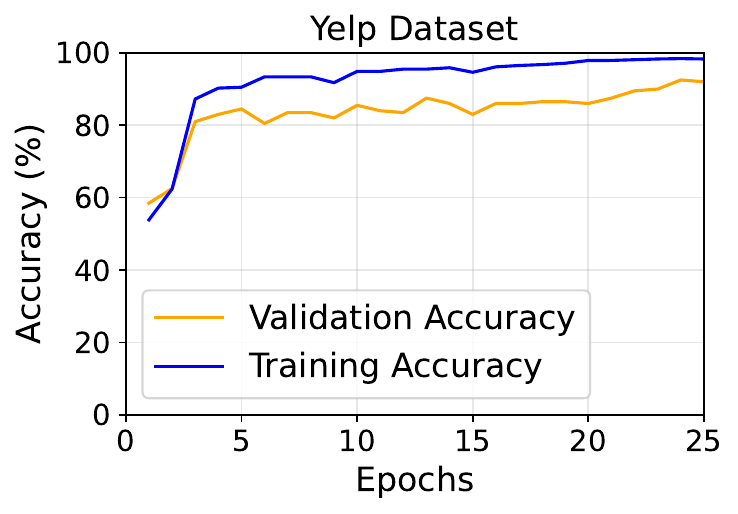}
                
            \end{subfigure}
            \hfill
            \begin{subfigure}[b]{0.33\textwidth}
                \includegraphics[width=\linewidth]{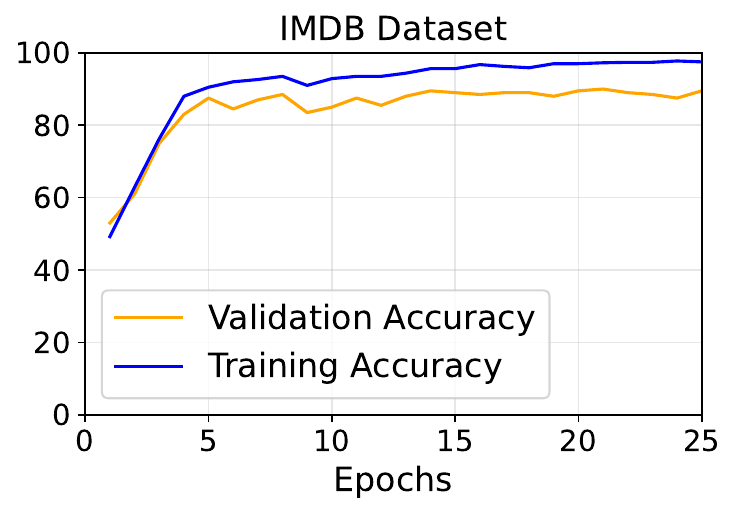}
               
            \end{subfigure}
            \hfill
            \begin{subfigure}[b]{0.33\textwidth}
                \includegraphics[width=\linewidth]{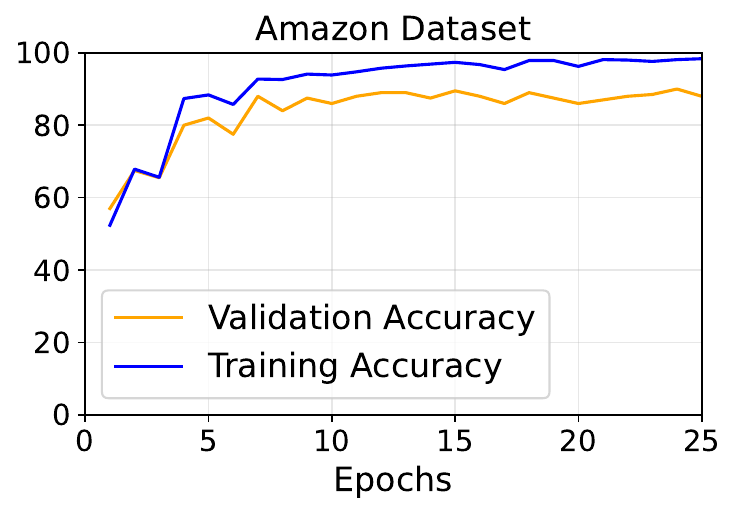}
            \end{subfigure}
           \captionsetup{skip=2pt} 
            \caption{Training and validation accuracy curves for the Yelp (\textit{left}), IMDB (\textit{middle}), and Amazon (\textit{right}) datasets in the sentiment classification task using the proposed QGT model. The consistent gap between training and validation curves indicates stable convergence and strong generalization across all three datasets.}
            \label{fig:model-accuracy-plots}
        \end{figure*}
    
    \subsection{QTransformerConv: Quantum Self-Attention}
    After encoding, each node feature is represented as a quantum state $\ket{\psi_i}$. We then apply a QTransformerConv layer, which performs message passing based on a quantum-en\-hanced self-attention mechanism. To extract query vectors for self-attention, we apply $U_Q(\boldsymbol{\theta})$ with $n$ qubits to each encoded state $\ket{\psi_i}$. As illustrated in Figure~\ref{fig:qgt-circuits} (right), the query circuit $U_Q(\boldsymbol{\theta})$ consists of three layers of single-qubit rotations---$R_x$, $R_y$, and $R_z$---applied to each qubit, followed by a ring-style CNOT entanglement layer. This is succeeded by an additional layer of $R_y$ rotations and a Quantum Fourier Transform (QFT) applied across all qubits. A similar circuit structure, $U_K(\boldsymbol{\phi})$, is used to generate key vectors, differing only in its parameter set $\boldsymbol{\phi}$. Both $U_Q(\boldsymbol{\theta})$ and $U_K(\boldsymbol{\phi})$ are fully differentiable and their parameters $\boldsymbol{\theta}$ and $\boldsymbol{\phi}$ are learned and optimized during training. The query $\mathbf{Q}_i$ and key $\mathbf{K}_i$ vectors are then obtained by repeatedly sampling from the circuit to measure the expectation values of some observables $\mathcal{O}_k$:
    \begin{align}
        \mathbf{Q}_i &= \left(\bra{\psi_i} U_Q^\dagger(\boldsymbol{\theta}) \mid \mathcal{O}_k \mid U_Q(\boldsymbol{\theta}) \ket{\psi_i}\right)_{k=0,\ldots,n-1} \\
        \mathbf{K}_i &= \left(\bra{\psi_i} U_K^\dagger(\boldsymbol{\phi}) \mid \mathcal{O}_k \mid U_K(\boldsymbol{\phi}) \ket{\psi_i}\right)_{k=0,\ldots,n-1}
    \end{align}

    The observables $\mathcal{O}_k$ can be any combination of Pauli-X, Pauli-Y and Pauli-Z measurements. In our proposed model, we apply a Pauli-X measurement on each qubit to get query $\mathbf{Q}_i$ and key $\mathbf{K}_i$ vectors. For a graph with $N$ nodes, this process is repeated independently for each node to obtain its corresponding query and key representations. Once the query $\mathbf{Q}_i$ and key $\mathbf{K}_i$ vectors are computed for all nodes, attention scores are calculated to determine the relevance of each neighboring node $j$ to a given node $i$. Similar to classical self-attention in  Equations~\eqref{classical-attn}~\&~\eqref{classical-attn-normalized}, the attention score between node $i$ and node $j$ is computed using a scaled dot-product, and then normalized via the softmax function:
    \begin{align}
        \alpha_{i,j} = \frac{\exp\left( \frac{\mathbf{Q}_i \cdot \mathbf{K}_j^T}{\sqrt{d}} \right)}{\sum_{m \in \mathcal{N}(i)} \exp\left( \frac{\mathbf{Q_i} \cdot \mathbf{K}_m^T}{\sqrt{d}} \right)}
        \label{eq:final_attn_scores}
    \end{align}
    This normalized score determines how much influence token $j$ exerts on token $i$ during the message-passing phase. Higher attention scores correspond to stronger influence from the source node. 
    
    \subsection{QTransformerConv: Message Passing}
    Once the quantum attention scores between token nodes are computed, the model proceeds with message passing over the graph to propagate contextual information. Quantum-enhanced attention values from neighboring tokens are aggregated to update each node's  embedding vector. Specifically, each token node sends messages to its directly connected neighbors (based on the edge set $E$), with each message weighted by the quantum attention scores $\alpha_{i,j}$ as defined in Equation~\ref{eq:final_attn_scores}. For each node $i$, the message received from a neighboring node $j \in \mathcal{N}(i)$ is computed as $m_{i,j} = \alpha_{i,j} \cdot \mathbf{V}_j$ where $\mathbf{V}_j$ is the projected feature vector (value vector) associated with node $j$. The total message aggregated at node $i$ from all of its neighbors is $ m_i = \sum_{j \in \mathcal{N}i} m_{i,j}$
    The node's feature vector is then updated by combining the original node representation $\mathbf{x}_i$ with the aggregated message, $\mathbf{x}_i^{'} = \mathbf{x}_i + m_i$. This update scheme preserves the identity of the original node while enriching its representation with context-aware features derived from its  neighborhood.

    \subsection{Classification Head}
   After the QTransformerConv layer refines the token (node) representations using quantum attention-guided message passing, the model aggregates the node-level information for sentence-level classification. A global mean pooling operation is applied across all node embeddings $\mathbf{x}'_i$. The pooled sentence representation $\mathbf{x}'_{\text{graph}}$ is computed as:
    \begin{align}
        \mathbf{x}'_{\text{graph}} = \frac{1}{N} \sum_{i=1}^{N} \mathbf{x}'_i
    \end{align}
    where $N$ is the number of nodes. This pooled representation is then passed through a fully connected layer to produce raw class logits $\mathbf{z}$, given by:
    \begin{align}
        \mathbf{z} = \mathbf{W}^T \mathbf{x}'_{\text{graph}} + \mathbf{b}
    \end{align}
    where $\mathbf{W}$ is the weight matrix and $\mathbf{b}$ is the bias. The logits $\mathbf{z}$ are used in the cross-entropy loss~\cite{pytorch_crossentropy_loss} function, which applies the softmax function internally to compute the probability distribution over classes. The loss is defined as:
    \begin{align}
    \mathcal{L} = - \sum_{i=1}^{c} y_i \log \left( \frac{\exp(z_i)}{\sum_{j=1}^{c} \exp(z_j)} \right)
    \end{align}
    where $c$ is number of classes, $y_i$ is the true label and $z_i$ is the logit for class $i$. During backpropagation, the model’s trainable parameters are updated to minimize this loss. The model parameters come from two main components: (1) the quantum self-attention mechanism, which includes $U_Q(\boldsymbol{\theta})$ and $U_K(\boldsymbol{\phi})$, each with $4 \times 4 = 16$ trainable parameters; and (2) the classical layers, including the embedding projection, value projection, and the classification head. Figure~\ref{fig:qgt-model} shows the trainable parameters in red. The detailed training setup and training parameters are discussed in Section~\ref{exp-setup}.

 \begin{figure}[t!]
        \centering
        
        \includegraphics[width=0.99\linewidth]{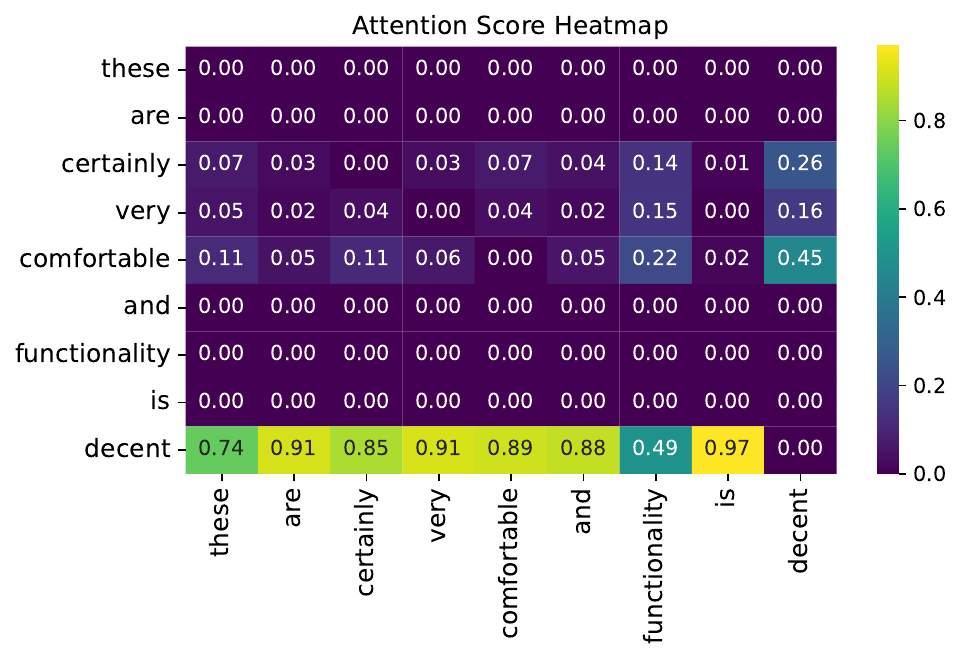}  
        \captionsetup{skip=2pt} 
        \caption{Attention score heatmap of a testing sample from the Amazon dataset. Lighter colors indicate higher attention given to the corresponding token from other tokens.}
        \label{fig:attn-score-amazon}
    \end{figure}

\section{Experimental Result \& Analysis}   
\label{exp-result}
 \begin{figure*}[t!]
        \includegraphics[width=0.595\textwidth]{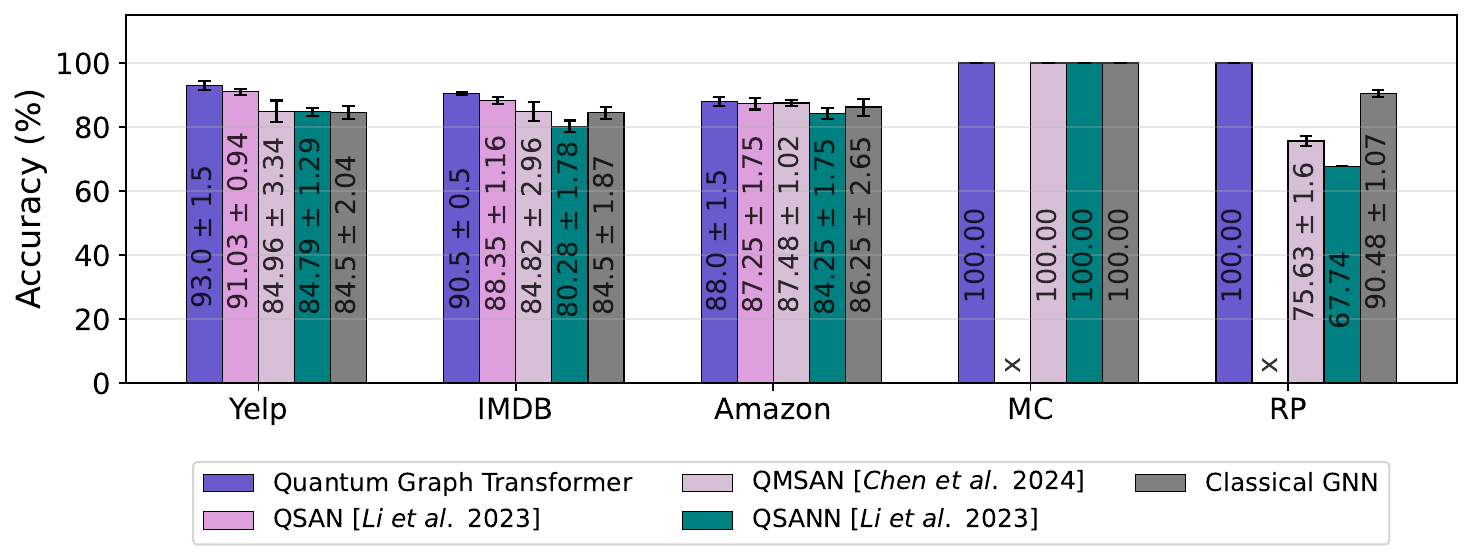}
        \includegraphics[width=0.4\textwidth]{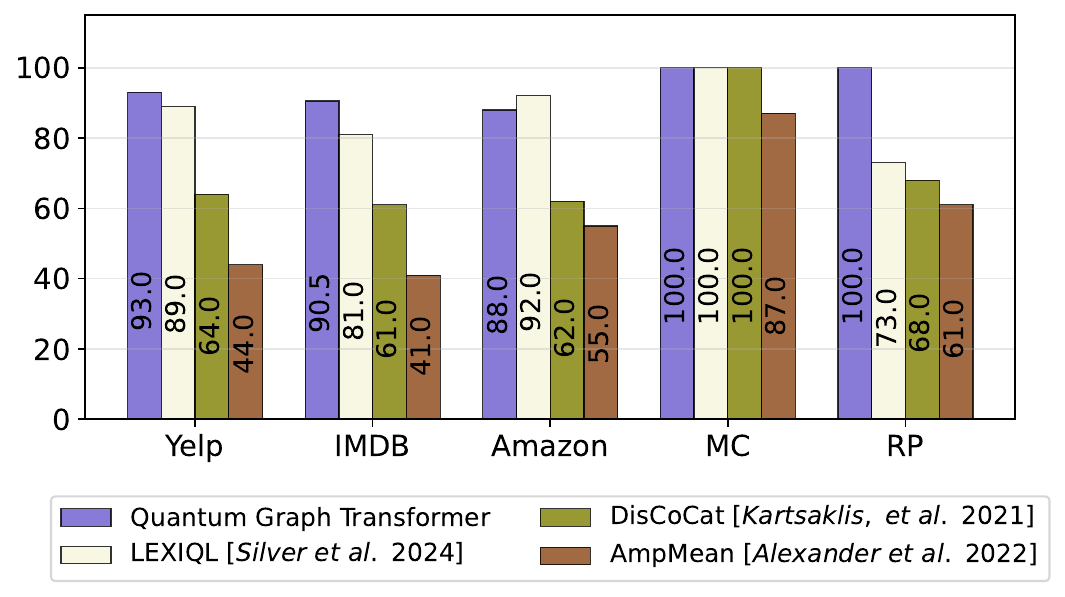}
        \captionsetup{skip=1pt} 
        \caption{(\textit{Left}) Test accuracy comparison between the proposed QGT model and existing attention-based models, including a classical graph transformer model. (\textit{Right}) Test accuracy comparison between the QGT model and other baseline models. The QGT model consistently outperforms all baselines, with performance comparable to LEXIQL~\cite{silver2024lexiql} on the Amazon dataset.}
        \label{fig:model-comp}
    \end{figure*}
\subsection{Experimental Setup} 
    \label{exp-setup}
    We evaluate the QGT model on five commonly used benchmark datasets for QNLP models. Three datasets are from Yelp, IMDB, and Amazon, consisting of $1,000$ samples each~\cite{kotzias2015group}. The Yelp dataset contains restaurant reviews with positive or negative sentiment labels, the IMDB dataset contains movie reviews with positive or negative sentiment labels, and the Amazon dataset includes user product review ratings from 1 to 5. The three datasets have maximum sentence lengths of $34$, $45$, and $30$, respectively. The other two datasets, MC (Meaning Classification) and RP (Relative Pronoun), are synthetic and generated as described in~\cite{lorenz2023qnlp}. They have relatively smaller vocabularies, and each sample has a length of 4 tokens. The MC dataset contains 130 samples, while the RP dataset has 105 samples. Each of the five datasets is split into 70\%, 10\%, and 20\% as training, validation, and testing sets, respectively, and a separate model is trained for each dataset. Training is performed using the PennyLane Lightning device~\cite{bergholm2018pennylane}, and the model is run on macOS with an Apple M3 Max chip and 64 GB of RAM.
   
    Each dataset is processed and trained using the QGT model described in Section~\ref{sec:QGT-Model}. In our experiments, we use $4$~qubits and a single layer of PQCs, $U_Q(\boldsymbol{\theta})$ and $U_K(\boldsymbol{\phi})$, to compute the query and key vectors, respectively. The parameters $\boldsymbol{\theta}$ and $\boldsymbol{\phi}$ are initialized from a normal distribution with mean $0$ and standard deviation $0.01$. We choose Adam optimizer with a learning rate of $0.01$, and employ cross-entropy loss~\cite{pytorch_crossentropy_loss}, which is well-suited for classification tasks involving categorical sentiment labels. We use a StepLR scheduler with step size of $5$ and decay factor (gamma) of $0.7$, a batch size of 32, and apply early stopping based on validation loss, with training running for up to 25 epochs. To encourage better exploration of the parameter space for $U_Q(\boldsymbol{\theta})$ and $U_K(\boldsymbol{\phi})$, we apply a reinforcement learning–based regularization strategy. The reward is defined as the negative training loss and is used to directly update the parameters of the PQCs. We also experimented with using multiple layers of $U_Q(\boldsymbol{\theta})$ and $U_K(\boldsymbol{\phi})$ as well as multi-headed attention. However, no significant improvements were observed on these datasets, likely due to their small size. 
    
    \subsection{Model Learning Dynamics}
    We first analyze the learning dynamics of the proposed QGT model to evaluate its performance. Figure~\ref{fig:model-learning-plots} presents the training loss curves for the Yelp, IMDB, and Amazon datasets. The plots show a consistent reduction in training loss over epochs, indicating effective learning across all three datasets. In each case, the loss decreases sharply during the initial epochs, followed by gradual convergence. Figure~\ref{fig:model-accuracy-plots} demonstrates the training and validation accuracy curves for the same datasets. The validation accuracy consistently follows the training accuracy,  showing the model’s ability to generalize while minimizing loss over epochs. Figure~\ref{fig:attn-score-amazon} illustrates the attention score heatmap of a test sample from the Amazon dataset, where lighter colors indicate higher attention weights. From the plot, we observe that tokens such as \textit{“decent”}, \textit{“comfortable”}, and \textit{“certainly”} receive higher attention scores from other tokens, as they contribute more to the positive sentiment label.

    \subsection{Comparison with existing Quantum NLP models}
    We compare the performance of the proposed QGT model with existing QNLP models developed for sentiment classification tasks. Figure~\ref{fig:model-comp} (\textit{left}) presents the test accuracy of the QGT model alongside other attention-based quantum NLP models (QSANN~\cite{li2024quantum}, QSAN~\cite{zhang2024light}, QMSAN~\cite{chen2025quantum}) evaluated on all five datasets: Yelp, IMDB, Amazon, MC, and RP. The plot also includes results from a classical graph transformer model, providing a comparative perspective against non-quantum architectures. The QGT model consistently achieves higher across all five datasets, demonstrating its effectiveness in capturing semantic structure through quantum graph-based attention mechanisms.  Figure~\ref{fig:model-comp} (\textit{right}) presents a performance comparison of the QGT model against several baseline models, including the grammar-aware DiscoCat~\cite{church2017word2vec, lorenz2023qnlp}, the ensemble-based LEXIQL model~\cite{silver2024lexiql}, and AmpMean~\cite{alexander2022quantum}. The plot demonstrates that the QGT model outperforms other baseline Quantum NLP techniques, with the exception of LEXIQL, on the Amazon dataset. Overall, the QGT model exhibits strong generalization performance across all baseline quantum NLP models and classical graph transformer models. 
     \begin{figure}[t!]
        \centering
        \vspace{-5px}
        \includegraphics[width=0.99\linewidth]{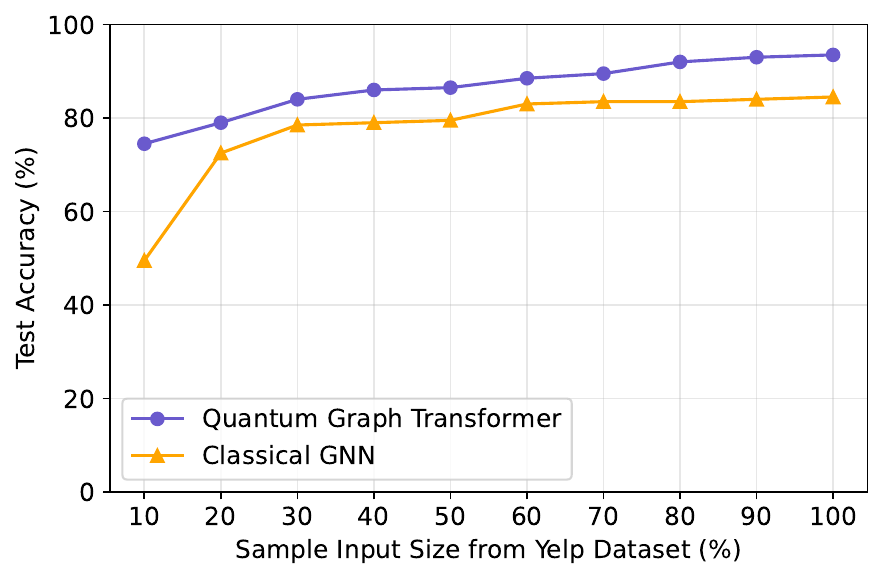}
        \captionsetup{skip=1pt} 
        \caption{Demonstrating the effect of varying sample sizes from the Yelp dataset on model learning performance. The plot shows accuracy on a fixed testing set for both the QGT model and a classical graph transformer. The QGT model demonstrates better learning efficiency with fewer samples.}
        \label{fig:sample-qgt-cgt}
    \end{figure}
    \subsection{Quantum NLP Advantage}
     In our proposed QGT model, the self-attention mechanism leverages PQCs to compute attention with significantly fewer parameters compared to the classical attention mechanism. While classical attention layers require high dimensional projections with large weight metrics, our quantum model generates query and key vectors  using only $32$ trainable parameters for each layers of $U_Q(\boldsymbol{\theta})$ and $U_K(\boldsymbol{\phi})$. Despite this reduced parameter count, the comparison in Figure~\ref{fig:model-comp} shows that the QGT model effectively learns attention and outperforms the classical graph transformer model. The key to this efficiency lies in the expressive power of PQCs, which enables the model to capture rich quantum representations within the Hilbert space.

     One of the major limitations of current classical NLP models is their reliance on large corpora of data and their computationally intensive architectures. Any viable alternative model should aim to overcome these constraints. To check sample efficiency, we vary the training data from 10\% to 100\% of the Yelp dataset while evaluating testing accuracy on a fixed test set. Figure~\ref{fig:sample-qgt-cgt} shows the testing accuracy performance of both the quantum and classical graph transformer models across varying training sample sizes. The plot show that the QGT model requires fewer training samples to achieve comparable performance. Specifically, the QGT model reaches 82\% accuracy using only 30\% of the training samples from Yelp dataset, whereas the classical graph transformer requires 60\% of the training samples to attain similar accuracy. This highlights the potential of quantum NLP models for data-efficient learning in NLP tasks.

\section{Limitations \& Future Work}
    The QGT model shows promising results for sentiment classification on small-scale datasets, outperforming existing QNLP techniques and classical graph transformers. However, its evaluation has been limited to small, synthetic datasets. Future work should explore QGT’s performance on larger datasets such as SST-5~\cite{socher2013recursive}, Twitter Sentiment~\cite{rosenthal2017semeval}, and the Kaggle IMDb Movie Review dataset~\cite{imdb_kaggle}. To adapt to these datasets, we may need to incorporate multiple layers of query and key PQCs and use multi-head quantum self-attention to capture richer semantic patterns. Each head in multi-head attention can be computed in parallel using separate quantum circuits, though the qubit count will grow linearly with the number of heads. Additionally, we observe improvements in parameter reduction and sample efficiency on small datasets. Future work could also investigate other quantum advantages such as generalization and resource requirements. Moreover, the performance of the  QGT model could be evaluated under the impact of noise. Additionally, the graph structure can be optimized by incorporating semantic or grammar-aware connections instead of fully connected token graphs.

\section{Conclusion}
    In this work, we propose the Quantum Graph Transformer (QGT) model for the sentiment classification task. The model introduces a novel QTransformerConv layer, which employs quantum self-attention for message passing between nodes. This architecture effectively captures the complex relationships between sentence tokens and their corresponding sentiment labels. Experimental results on benchmark datasets demonstrate the potential of graph-based quantum NLP models in learning and performing NLP tasks.

\section{Acknowledgements}
The research presented in this article was supported by
the NNSA’s Advanced Simulation and Computing Beyond Moore’s Law program at Los Alamos National Laboratory and
the Laboratory Directed Research and Development program of Los Alamos National Laboratory under project number 20230049DR. 
This work has been assigned LANL technical report number LA-UR-25-24685.

\bibliographystyle{ACM-Reference-Format}
\bibliography{qgt-nlp-augmented}

\end{document}